\documentclass[letterpaper]{article}
\usepackage{times}
\usepackage{microtype}
\usepackage{latexsym} 
\usepackage{theorem}
\usepackage{xspace}
\usepackage{amsmath}
\usepackage{mathtools}
\usepackage{amssymb}
\usepackage{booktabs}
\usepackage{paralist}
\usepackage{multirow}

\newtheorem{proposition}{Proposition}
\newtheorem{definition}{Definition}
\newtheorem{theorem}{Theorem}

\newtheorem{remark}{Remark}
\newtheorem{example}{Example}

\newcommand{\goal}{G}
\newcommand{\game}{P}
\newcommand{\states}{S}
\newcommand{\memstates}{M}
\newcommand{\act}{\mathcal{A}}
\newcommand{\trans}{\delta}
\newcommand{\obs}{\mathcal{Z}}
\newcommand{\obsmap}{\mathcal{O}}
\newcommand{\distr}{\mathcal{D}}
\newcommand{\initd}{I}

\newcommand{\last}{\mathsf{Last}}
\newcommand{\Cone}{\mathsf{Cone}}
\newcommand{\Reach}{\mathsf{Reach}}
\newcommand{\straa}{\sigma}
\newcommand{\prb}{\mathbb{P}}
\newcommand{\msize}{\mu}
\newcommand{\powerset}{\mathcal{P}}

\usepackage{nicefrac}
\usepackage{tikz}
\usetikzlibrary{fit}
\usetikzlibrary{backgrounds}
\usetikzlibrary{automata}
\usetikzlibrary{shapes}
\usetikzlibrary{matrix}
\usetikzlibrary{fit}
\usetikzlibrary{calc}
\usetikzlibrary{positioning}

\tikzstyle{Player1}=[circle, thick, minimum size=0.6cm, inner sep=0cm,draw=black]
\tikzstyle{State}=[circle, thick, minimum size=0.6cm, inner sep=0cm,draw=black]
\tikzstyle{Final}=[circle, accepting, thick, minimum size=0.6cm, inner sep=0cm,draw=black]
\tikzstyle{RState}=[circle, very thick, minimum size=0.8cm, inner sep=0cm,draw=red]

\title{A Symbolic SAT-based Algorithm for Almost-sure Reachability with Small Strategies in POMDPs}
\author{Krishnendu Chatterjee\\IST Austria \and Martin Chmel\'ik\\IST Austria \and Jessica Davies\\IST Austria}
\date{}
\begin{document}

\maketitle

\begin{abstract}
POMDPs are standard models for probabilistic planning problems, where an agent interacts with an uncertain environment. We study the problem of almost-sure reachability, where given a set of target states, the question is to decide whether there is a policy to ensure that the target set is reached with probability 1 (almost-surely). While in general the problem is EXPTIME-complete, in many practical cases policies with a small amount of memory suffice. Moreover, the existing solution to the problem is explicit, which first requires to construct explicitly an exponential reduction to a belief-support MDP. In this work, we first study the existence of observation-stationary strategies, which is NP-complete, and then small-memory strategies. We present a symbolic algorithm by an efficient encoding to SAT and using a SAT solver  for the problem. We report experimental results demonstrating the scalability of our symbolic (SAT-based) approach.
\end{abstract}

\section{Introduction}

The de facto model for dynamic systems with probabilistic and nondeterministic
behavior are \emph{Markov decision processes (MDPs)}~\cite{Howard}.
MDPs provide the appropriate model to solve control and probabilistic planning 
problems~\cite{FV97,Puterman}, where the nondeterminism represents 
the choice of the control actions for the controller (or planner), 
while the stochastic response of the system to control actions
is represented by the probabilistic behavior. 
In \emph{perfect-observation (or perfect-information) MDPs}, to resolve
the nondeterministic choices among control actions the controller 
observes the current state of the system precisely, 
whereas in \emph{partially observable MDPs (POMDPs)} the 
state space is partitioned according to observations that the controller 
can observe, i.e., the controller can only view the observation 
of the current state (the partition the state belongs to), but not the precise 
state~\cite{PT87}.
POMDPs are widely used in several applications, such as in computational 
biology~\cite{Bio-Book}, speech processing~\cite{Mohri97}, 
image processing~\cite{IM-Book}, software verification~\cite{CCHRS11}, 
robot planning~\cite{KGFP09,kaelbling1998planning}, 
reinforcement learning~\cite{LearningSurvey}, to name a few.

\smallskip\noindent{\em Reachability objectives and their computational problems.}
We consider POMDPs with one of the most basic and fundamental objectives,
namely, \emph{reachability objectives}. 
Given a set of target states, the reachability objective requires that some 
state in the target set is visited at least once. 
The main computational problems for POMDPs with reachability objectives 
are as follows: (a)~the \emph{quantitative} problem asks for the existence 
of a policy (that resolves the choice of control actions) that ensures
the reachability objective with probability at least $0<\lambda\leq 1$; 
and (b)~the \emph{qualitative} problem is the special case of the 
quantitative problem with $\lambda=1$ (i.e., it asks that the objective
is satisfied almost-surely).

\smallskip\noindent{\em Significance of qualitative problems.}
The qualitative problem is of great importance as in several applications it 
is required that the correct behavior happens with probability~1, e.g., 
in the analysis of randomized embedded schedulers, the important question is 
whether every thread progresses with probability~1.
Also in applications where it might be sufficient that the correct behavior 
happens with probability at least $\lambda<1$, the correct choice of 
the threshold $\lambda$ can be still challenging, due to simplifications and 
imprecisions introduced during modeling.  For example, in the analysis of randomized 
distributed algorithms it is common to require correctness with probability~1 
(e.g.,~\cite{PSL00}). 
Finally, it has been shown recently~\cite{CCGK15} that for the important 
problem of minimizing the total expected cost to reach the target 
set~\cite{Bertsekas95,BG09,KMWG11} (under positive cost functions), it  
suffices to first compute the almost-sure winning set, and then apply any 
finite-horizon algorithm for approximation.
Besides its importance in practical applications, almost-sure convergence, 
like convergence in expectation, is a fundamental concept in probability theory,
and provides the strongest probabilistic guarantee~\cite{Durrett}.

\smallskip\noindent{\em Previous results.} 
The quantitative analysis problem for POMDPs with reachability objectives 
is undecidable~\cite{PazBook} (and the undecidability result even holds for any 
approximation~\cite{MHC03}).
In contrast, the qualitative analysis problem is 
EXPTIME-complete~\cite{CDH10a,BGB12}.
The main algorithmic idea to solve the qualitative problem (that originates 
from~\cite{CDHR06}) is as follows: first construct the belief-support MDP 
explicitly (which is an exponential size perfect-information MDP where every 
state is the support of a belief), and then solve the qualitative analysis on 
the perfect-information MDP. 
Solving the qualitative analysis problem on the
resulting MDP can be done using any one of several known polynomial-time
algorithms, which are based on discrete graph theoretic approaches~\cite{CJH03,CH14,CH11}.
This yields the EXPTIME upper bound for the qualitative
analysis of POMDPs, and the EXPTIME lower bound has been established in~\cite{CDH10a}.

\smallskip\noindent{\em Drawbacks.}
There are two major drawbacks of the present solution for the qualitative 
problem for POMDPs with reachability objectives.
First, the algorithm requires to explicitly construct an exponential-size MDP, 
and there is no symbolic algorithm (that avoids the explicit construction) 
for the problem. 
Second, even though in practice a small amount of memory in policies might 
suffice, the construction of the belief-support MDP always searches for an 
exponential size policy (which is only required in the worst case). 
There is no algorithmic approach 
for small-memory policies for the problem.

\smallskip\noindent{\em Our contributions.}
In this work our main contributions are as follows.
First, we consider the qualitative analysis problem with respect to the special case of \emph{observation-stationary}  
 (i.e., memoryless) policies.
This problem is NP-complete. Motivated by the impressive performance of state-of-the-art SAT solvers in applications from AI as well as many other fields~\cite{lingeling,rintanen11,BCCFZ99}, we present an efficient reduction of our problem to 
SAT. This results in a practical, symbolic algorithm for the 
almost-sure reachability problem in POMDPs.  
We then show how our encoding to SAT can be extended to search for policies 
that use only a small amount of memory. 
Thus we present a symbolic SAT-based algorithm that determines the existence of 
small-memory policies in POMDPs that can ensure that a target set is reached 
almost-surely.
Our encoding is efficient: in the worst case it uses a quadratic number of variables 
and a cubic number of clauses, as compared to a naive encoding that uses a quartic 
(fourth power) number of clauses; and in practice our encoding uses just a linear 
number of variables and a quadratic number of clauses. 
Moreover, our encoding is incremental (it incrementally searches over lengths 
of paths), which may be further exploited by incremental SAT solvers
(see Remark~\ref{remark} for details). 
An important consequence of our result is that any improvement in SAT-solvers 
(improved solvers or parallel solvers), which is an active research area, 
carries over to the qualitative problem for POMDPs.
We have implemented our approach and our experimental results show that our 
approach scales much better, and can solve large POMDP instances where the 
previous method fails.

\smallskip\noindent{\em Comparison with contingent or strong planning.}
We consider the qualitative analysis problem which is different as compared 
to strong or contingent planning~\cite{MBKS14,CPRT03,APG09}.
The strong planning problem has been also considered under partial observation 
in~\cite{BCRT06,R04,BPG09}.
The key difference of strong planning and qualitative analysis is as follows:
in contingent planning the probabilistic aspect is treated as an adversary, 
whereas in qualitative analysis though the precise probabilities do not 
matter, still the probabilistic aspect needs to be considered. 
For a detailed discussion with illustrative examples see 
Appendix~\ref{sec:app:cont}.

\smallskip\noindent{\em Comparison with strong cyclic planning.} 
The qualitative analysis problem is equivalent to the strong cyclic planning 
problem.
The strong cyclic problem was studied in the perfect information 
setting in~\cite{CPRT03} and later extended to the partial information 
setting in~\cite{BCP06}. 
However, there are two crucial differences of our work wrt~\cite{BCP06}: 
(i)~We consider the problem of finding small strategies as compared to general 
strategies. 
We show that our problem is NP-complete. In contrast, it is known that the 
qualitative analysis problem for POMDPs with general strategies is 
EXPTIME-complete~\cite{CDH10a,BGB12}. 
Thus the strong cyclic planning with general strategies considered 
in~\cite{BCP06} is also EXPTIME-complete, whereas our problem is NP-complete.
Thus there is a significant difference in the complexity of the problem 
we consider.
(ii)~The work of~\cite{BCP06} presents a BDD-based implementation, whereas we 
present a SAT-based implementation. Note that since~\cite{BCP06} considers 
an EXPTIME-complete problem in general there is no efficient reduction to SAT.
(iii)~Finally, the equivalence of strong cyclic planning and qualitative 
analysis of POMDPs imply that our results present an efficient SAT-based 
implementation to obtain small strategies in strong cyclic planning
(also see Appendix~\ref{sec:app:cont2} for a detailed discussion).

\section{Preliminaries}
\begin{definition}\textbf{POMDPs.}
A \emph{Partially Observable Markov Decision Process (POMDP)} is defined as a
tuple $\game=(\states,\act,\trans,\obs,\obsmap,\initd)$ where:
\begin{compactitem}
\item (i)~$\states$ is a finite set of states;
\item (ii)~$\act$ is a finite alphabet of \emph{actions};
\item (iii)~$\trans:S\times\act \rightarrow \distr(S)$ is a 
 \emph{probabilistic transition function} that given a state $s$ and an
 action $a \in \act$ gives the probability distribution over the successor 
 states, i.e., $\trans(s,a)(s')$ denotes the transition probability from 
 $s$ to $s'$ given action~$a$; 
 \item (iv)~$\obs$ is a finite set of \emph{observations}; 
 \item (v)~$\initd \in \states$ is the unique initial state;
 \item (vi)~$\obsmap:\states\rightarrow \obs$ is an \emph{observation function} that 
  maps every state to an observation. For simplicity w.l.o.g. we consider 
  that $\obsmap$ is a deterministic function (see~\cite[Remark~1]{CCGK14}).
\end{compactitem}
\end{definition}

\noindent \textbf{Plays and Cones.}
A \emph{play} (or a path) in a POMDP is an infinite sequence $(s_0,a_0,s_1,a_1,s_2,a_2,\ldots)$ 
of states and actions such that $s_0 = \initd$ and for all $i \geq 0$ we have $\trans(s_i,a_i)(s_{i+1})>0$.
We write $\Omega$ for the set of all plays.
For a finite prefix $w \in ( \states \cdot \act)^* \cdot \states$ of a play, we denote by $\Cone(w)$ the 
set of plays with $w$ as the prefix (i.e., the cone or cylinder of the prefix $w$), 
and denote by $\last(w)$ the last state of $w$.
For a finite prefix $w=(s_0,a_0,s_1,a_1,\ldots,s_n)$ 
we denote by 
$\obsmap(w)=(\obsmap(s_0),a_0,\obsmap(s_1),a_1,\ldots,\obsmap(s_n))$ 
the observation and action sequence associated with $w$.

\smallskip\noindent \textbf{Strategies (or policies).}
A \emph{strategy (or a policy)} is a recipe to extend prefixes of plays and 
is a function $\straa: ( \states \cdot \act)^* \cdot \states \to \distr(A)$ that given a finite 
history (i.e., a finite prefix of a play) selects a probability distribution 
over the actions.
Since we consider POMDPs, strategies are \emph{observation-based}, i.e., 
for all histories $w=(s_0,a_0,s_1,a_1,\ldots,a_{n-1},s_n)$ and 
$w'=(s_0',a_0,s_1',a_1,\ldots,a_{n-1},s_n')$ such that for all 
$0\leq i \leq n$ we have $\obsmap(s_i)=\obsmap(s_i')$ (i.e., $\obsmap(w) = \obsmap(w')$), we must have 
$\straa(w)=\straa(w')$.
In other words, if the observation sequence is the same, then the strategy 
cannot distinguish between the prefixes and must play the same. Equivalently, we can define a POMDP strategy as a function $\straa: (\obs \cdot \act)^* \cdot \obs \rightarrow \distr(\act)$.

\smallskip\noindent \textbf{Observation-Stationary (Memoryless) Strategies.}
A strategy $\straa$ is observation-stationary (or \emph{memoryless}) if it depends only on the current observation, i.e., 
whenever for two histories $w$ and $w'$, we have
$\obsmap(\last(w)) = \obsmap(\last(w'))$, then $\straa(w) = \straa(w')$. Therefore, a memoryless strategy is just a mapping from observations to a distribution over actions: $\straa : \obs \rightarrow \distr(A)$. We may also define a memoryless strategy as a mapping from states to distributions over actions (i.e., $\straa : \states \rightarrow \distr(\act)$), as long as $\straa(s) = \straa(s')$ for all states $s,s'\in \states$ such that $\obsmap(s) = \obsmap(s')$. All three definitions are equivalent, so we will use whichever definition is most intuitive.
We define the set of states that can be reached using a memoryless strategy recursively: $\initd \in R_\straa$, and if $s \in R_\straa$ then $s' \in R_\straa$ for all $s'$ such that there exists an action $a$ where $\trans(s, a)(s') > 0$ and $\straa(s)(a) > 0$.
Let $\pi_k(s,s') = (s_1,a_1,...,s_k)$ be a path of length $k$ from $s_1 = s$ to $s_k = s'$. We say that $\pi_k(s,s')$ is \emph{compatible} with $\straa$ if $\straa(s_i)(a_i) > 0$ for all $1\leq i < k$.

\smallskip\noindent \textbf{Strategies with Memory.}
A strategy with memory is a tuple $\straa = (\straa_u, \straa_n, \memstates, m_0)$ where: (i) $\memstates$ is a finite set of \emph{memory states}. (ii) The function $\straa_n : \memstates \rightarrow \distr(\act)$ is the \emph{action selection function} that given the current memory state gives the probability distribution over actions. (iii) The function $\straa_u : \memstates \times \obs \times \act \rightarrow \distr(\memstates)$ is the \emph{memory update function} that given the current memory state, the current observation and action, updates the memory state probabilistically. 
(iv)~The memory state $m_0 \in \memstates$ is the \emph{initial memory state}.

\smallskip\noindent \textbf{Probability Measure.}
Given a strategy $\straa$ and a starting state~$\initd$, the unique probability measure obtained given 
$\straa$ is denoted as $\prb_\initd^{\straa}(\cdot)$.
We first define a measure $\rho_\initd^\straa(\cdot)$ on cones.
For $w=\initd$ we have $\rho_\initd^\straa(\Cone(w))=1$, and 
for $w=s'$ where $ \initd \neq s'$ we have  $\rho_\initd^\straa(\Cone(w))=0$; and 
for $w' = w \cdot a\cdot s$ 
we have 
$\rho_\initd^\straa(\Cone(w'))= \rho_\initd^\straa(\Cone(w)) \cdot \straa(w)(a) \cdot \trans(\last(w),a)(s)$. 
By Carath\'eodory's extension theorem, the function $\rho_\initd^\straa(\cdot)$
can be uniquely extended to a probability measure $\prb_\initd^{\straa}(\cdot)$
over Borel sets of infinite plays~\cite{Billingsley}.

Given a set of target states, the reachability objective requires that a target state is visited at least once.
\begin{definition}\textbf{Reachability Objective.}
Given a set $T \subseteq \states$ of target states, the reachability objective 
is $\Reach(T) = \{(s_0,a_0,s_1,a_1,...) \in \Omega | \exists i\geq 0 : s_i \in T \}$. \end{definition}
 
In the remainder of the paper, we assume that the set of target states contains a single \emph{goal} state, i.e., $T = \{ \goal\} \subseteq \states$. We can assume this w.l.o.g. because it is always possible to add an additional state $\goal$ with transitions from all target states in $T$ to $\goal$.
\begin{definition}\textbf{Almost-Sure Winning.}
Given a POMDP $\game$ and a reachability objective $\Reach(T)$, a strategy $\straa$ is \emph{almost-sure winning} iff $\prb_\initd^\straa(\Reach(T)) = 1$.  
\end{definition}
In the sequel, whenever we refer to a winning strategy, we mean an almost-sure winning strategy.

\section{Almost-Sure Reachability with Memoryless Strategies}

In this section we present our results concerning the complexity of almost-sure reachability with memoryless strategies. First, we show that memoryless strategies for almost-sure reachability take a simple form. The following proposition states that it does not matter with which positive probability an action is played. 

\begin{proposition}
\label{prop:uniform}
A POMDP $\game$ with a reachability objective $\Reach(T)$ has a memoryless winning strategy if and only if it has a memoryless winning strategy $\straa$ such that for all $a,a'\in \act$ and $s\in \states$, if $\straa(s)(a) > 0$ and $\straa(s)(a') > 0$ then $\straa(s)(a) = \straa(s)(a')$.  
\end{proposition}
Intuitively, $\straa$ only distinguishes between actions that must not be played, and therefore have probability $0$, and those that may be played (having  probabilities $> 0$). This proposition implies that we do not need to determine precise values for the positive probabilities when designing a winning strategy. In the following, we will for simplicity slightly abuse terminology: when we refer to a strategy or distribution as being \emph{uniform}, we actually mean a distribution of this type.  

The following result shows that determining whether there is a memoryless winning strategy reduces to finding finite paths from states to the target set.  
\begin{proposition}
\label{prop:path}
A memoryless strategy $\straa$ is a winning strategy if and only if for each state $s\in R_\straa$, there is a path $\pi_k(s, G)$ compatible with $\straa$, for some finite $k \leq |\states|$.   
\end{proposition}

Intuitively, the strategy must prevent the agent from reaching a state from which the target states can not be reached. 
It follows that determining whether there exists a memoryless, almost-sure winning strategy is in the complexity class NP.
An NP-hardness result was established for a similar problem, namely, 
memoryless strategies in two-player games with partial-observation, in~\cite[Lemma~1]{CKS13}.
The reduction constructed a game that is a DAG (directed acyclic graph), 
and replacing the adversarial player with a uniform distribution over choices shows 
that the almost-sure reachability problem under memoryless strategies in POMDPs is 
also NP-hard.

\begin{theorem}
\label{thm:nphard}
The problem of determining whether there exists a memoryless, almost-sure winning strategy for a POMDP $\game$ and reachability objective $\Reach(T)$ is NP-complete.
\end{theorem}

The complexity of the almost-sure reachability problem for memoryless strategies suggests a possible approach to solve this problem in practice. We propose to find a memoryless winning strategy by encoding the problem as an instance of SAT, and then executing a state-of-the-art SAT solver to find a satisfying assignment or prove that no memoryless winning strategy exists.   
 
\subsection{SAT Encoding for Memoryless Strategies}
\label{sec:encmemoryless}

Next, we show how to encode the almost-sure reachability problem for memoryless strategies as a SAT problem. We will define a propositional formula $\Phi_k$ for an integer parameter $k\in \mathbb{N}$, in Conjunctive Normal Form, such that $\Phi_k$ (for a sufficiently large $k$) is satisfiable if and only if the POMDP $\game$ has a memoryless, almost-sure winning strategy for reachability objective $\Reach(\goal)$.

By Propositions~\ref{prop:uniform} and~\ref{prop:path}, we seek a function from states to subsets of actions, $\straa : \states \rightarrow \powerset(\act)$ (where $\powerset(\act)$ is the powerset of actions) such that for each state $s \in R_\straa$, there is a path $\pi_k(s,G)$ compatible with $\straa$ for some $k \leq |\states|$.
The value of $k$ will be a parameter of the SAT encoding. If we take $k$ to be sufficiently large, e.g., $k = |\states|$ then one call to the SAT solver will be sufficient to determine if there exists a winning strategy. If $k = |\states|$ and the SAT solver determines that $\Phi_k$ is unsatisfiable, it will imply that there is no memoryless winning strategy.

We describe the CNF formula $\Phi_k$ by first defining all of its Boolean variables, followed by the clausal constraints over those variables.

\smallskip\noindent \textbf{Boolean Variables.}
The Boolean variables of $\Phi_k$ belong to three groups, which are defined as follows:
\begin{compactenum}
\item $\{A_{ij}\}, 1 \leq i \leq |\states|$, $1 \leq j \leq |\act|$. The Boolean variable $A_{ij}$ is the proposition that the probability of playing action $j$ in state $i$ is greater than zero, i.e., that $\straa(i)(j) > 0$.

\item $\{C_{i}\}, 1\leq i \leq |\states|$. The Boolean variable $C_i$ is the proposition that state $i$ is reachable using $\straa$ (i.e, these variables define $R_\straa$).

\item $\{P_{ij} \}, 1 \leq i \leq |\states|, 0 \leq j \leq k$. The Boolean variable $P_{ij}$ represents the proposition that from state $i\in \states$ there is a path to the goal of length \emph{at most $j$}, that is compatible with the strategy.
\end{compactenum}

\smallskip\noindent \textbf{Logical Constraints.} The following clause is defined for each $i \in \states$, to ensure that at least one action is chosen in each state:
\begin{displaymath}
\bigvee_{j \in \act} A_{ij}
\end{displaymath}

To ensure that the strategy is observation-based, it is necessary to ensure that if two states have the same observations, then the strategy behaves identically. This is achieved by adding the following constraint for all pairs of states $i\neq j$ such that $\obsmap(i) = \obsmap(j)$, and all actions $r \in \act$:
\begin{displaymath}
A_{ir} \iff A_{jr}
\end{displaymath}

The following clauses ensure that the $\{C_i\}$ variables will be assigned True, for all states $i$ that are reachable using the strategy defined by the $\{A_{ij}\}$ variables:  
\begin{displaymath}
\neg C_i \vee \neg A_{ij} \vee C_{\ell}
\end{displaymath}
Such a clause is defined for each pair of states $i,\ell \in \states$ and action $j \in \act$ for which $\trans(i,j)(\ell) > 0$. Furthermore, the initial state is reachable by the strategy, which is expressed by adding the single clause:
\begin{displaymath}
C_I
\end{displaymath} 

We introduce the following unit clauses, which say that from the goal state, the goal state is reachable using a path of length at most $0$:
\begin{displaymath}
\left(P_{G,j}\right) \textrm{ for all } 0\leq j \leq k
\end{displaymath} 

For each state $i\in \states$, we introduce the following clause that ensures if $i$ is reachable, then there is a path from $i$ to the goal that is compatible with the strategy.  
\begin{displaymath}
\left(\neg C_i \vee P_{ik}\right) 
\end{displaymath}

Finally, we use the following constraints to define the value of the $\{P_{ij}\}$ variables in terms of the chosen strategy.
\begin{displaymath}
P_{ij} \iff \bigvee_{a \in \act}\left[A_{ia} \wedge \left(\bigvee_{i' \in \states : \trans(i,a)(i') > 0}P_{i',j -1} \right)\right]
\end{displaymath}

This constraint is defined for each $i\in \states$, and $1\leq j \leq k$. We translate this constraint to clauses using the standard Tseitin encoding~\cite{tseitin}, which introduces additional variables in order to keep the size of the clausal encoding linear. 

The conjunction of all clauses defined above forms the CNF formula $\Phi_k$. 

\begin{theorem}
If $\Phi_k$ is satisfiable, for any $k$, then a memoryless winning strategy $\straa : \states \rightarrow \distr(\act)$ can be extracted from the truth assignment to the variables $\{A_{ij}\}$. If $\Phi_k$ is unsatisfiable for $k = |\states|$ then there is no memoryless winning strategy.
\end{theorem}

The number of variables in $\Phi_k$ is $O(|\states| \cdot |\act| + |\states|\cdot k)$, and the number of clauses is $O(|\states|^2\cdot |\act|\cdot k)$. Note that the number of actions, $|\act|$, is usually a small constant, while the size of the state space, $|\states|$, is typically large. The number of variables is quadratic in the size of the state space, while the number of clauses is cubic (recall $k \leq |\states|$). 

\begin{remark}
\label{remark} 
\normalfont A naive SAT encoding would introduce a Boolean variable $X_{ij\ell}$ for each $i,\ell \in \states$, $1\leq j \leq k$, to represent the proposition that the $j^{th}$ state along a path from state $i$ to the goal is $\ell \in \states$. However, using such variables to enforce the existence of paths from every reachable state to the goal, instead of the variables $\{P_{ij}\}$ which we used above, results in a formula with a cubic number of variables and a quartic (fourth power) number of clauses. Thus our encoding has the theoretical advantage of being considerably smaller than the naive encoding.
Our encoding also offers two main practical advantages. First, it is possible to find a winning strategy, if one exists, using $k \ll |\states|$, by first generating $\Phi_k$ for small values of $k$. If the SAT solver finds $\Phi_k$ to be unsatisfiable, then we can increase the value of $k$ and try again.  Otherwise, if the formula is satisfiable, we have found a winning strategy and we can stop immediately. In this way, we are usually able to find a memoryless winning strategy (if one exists) very quickly, using only small values of $k$. So in practice, the size of $\Phi_k$ is actually only quadratic in $|\states|$.  Second, our encoding allows to take advantage of SAT solvers that offer an incremental interface, which supports the addition and removal of clauses between calls to the solver (though this is not exploited in our experimental results). 
\end{remark}

\section{Almost-Sure Reachability with Small-Memory Strategies}
For some POMDPs, a memoryless strategy that wins almost-surely may not exist. However, in some cases giving the agent a small amount of memory may help. We extend our SAT approach to the case of small-memory strategies in this section.

\begin{definition}
A small-memory strategy is a strategy with memory, $\straa = (\straa_u, \straa_n, \memstates, m_0)$, such that $|\memstates| = \msize$ for some small constant $\msize$.
\end{definition}
We will refer to the number of memory states, $\msize$, as the \emph{size} of the small-memory strategy. 
Propositions~\ref{prop:uniform} and~\ref{prop:path} and Theorem~\ref{thm:nphard} carry over to the case of small-memory strategies.

\begin{proposition}
A POMDP $\game$ with reachability objective $\Reach(T)$ has a small-memory winning strategy of size $\msize$ if and only if it has a small-memory winning strategy of size $\msize$ where both the action selection function and the memory update function are uniform.
\end{proposition}

We must modify the definition of a compatible path in the case of small-memory strategies, to also keep track of the sequence of memory states. Let $\pi_k(s,m,s',m') = (s_1,m_1,a_1,...,s_k,m_k,a_k)$ be a finite sequence where $s = s_1$, $m = m_1$, $s' = s_k$ and $m' = m_k$, and for all $1 \leq i \leq k$, $s_i \in \states$, $m_i \in \memstates$ and $a_i \in \act$. Then we say that $\pi_k(s,m,s',m')$ is a path compatible with small-memory strategy $\straa$ if for all $1 \leq i < k$, we have $\trans(s_i, a_i)(s_{i+1}) > 0$, $\straa_n(m_i)(a_i) > 0$, and $\straa_u(m_i,\obsmap(s_i),a_i)(m_{i+1}) > 0$. 

 Let $R_\straa$ be the set of all pairs $(s, m) \in \states \times \memstates$ such that there exists a finite-length path $\pi_k(I,m_0,s,m)$ that is compatible with $\straa$.

\begin{proposition}
A small-memory strategy $\straa$ is winning if and only if for each $(s,m) \in R_\straa$ there is a path $\pi_k(s,m,G,m')$ for some $k \leq |\states|\cdot |\memstates|$ and some $m' \in \memstates$, that is compatible with $\straa$. 
\end{proposition}

\begin{theorem}\label{thm:small}
The problem of determining whether there exists a winning, small-memory strategy of size $\msize$, where $\msize$ is a constant, is NP-complete.
\end{theorem}

Therefore, we may also find small-memory winning strategies using a SAT-based approach.
We remark that Theorem~\ref{thm:small} holds even if $\msize$ is polynomial
in the size of the input POMDP. 

\subsection{SAT Encoding for Small-Memory Strategies}
\label{sec:encsmall}

The SAT encoding from Section~\ref{sec:encmemoryless} can be adapted for the purpose of finding small-memory winning strategies. Given a POMDP $\game$, reachability objective $\Reach(G)$, a finite set of memory states $\memstates$ of size $\msize$, an initial memory state $m_0 \in \memstates$, and a path length $k \leq |\states|\cdot |\memstates|$, we define the CNF formula $\Phi_{k,\msize}$ as follows.

\smallskip\noindent \textbf{Boolean Variables.} 
We begin by defining variables to encode the action selection function $\straa_n$. We introduce a Boolean variable $A_{ma}$ for each memory-state $m \in \memstates$ and action $a\in \act$, to represent that action $a$ is among the possible actions that can be played by the strategy, given that the memory-state is $m$, i.e., that $\straa_n(m)(a) > 0$. 

The next set of Boolean variables encodes the memory update function. We introduce a Boolean variable $M_{m, z, a, m'}$ for each pair of memory-states $m,m' \in \memstates$, observation $z\in \obs$ and action $a \in \act$. If such a variable is assigned to True, it indicates that if the current memory-state is $m$, the current observation is $z$, and action $a$ is played, then it is possible that the new memory-state is $m'$, i.e., $\straa_u(m,z,a)(m') > 0$.

Similarly to the memoryless case, we also introduce the Boolean variables $C_{i, m}$ for each state $i \in \states$ and memory state $m \in \memstates$, that indicate which (state, memory-state) pairs are reachable by the strategy.

We define variables $\{P_{i,m,j}\}$ for all $i \in \states$, $m \in \memstates$, and $0 \leq j \leq k$, similarly to the memoryless case. The variable $P_{i,m,j}$ corresponds to the proposition that there is a path of length at most $j$ from $(i,m)$ to the goal, that is compatible with the strategy.
 
\smallskip\noindent \textbf{Logical Constraints.} We introduce the following clause for each $m \in \memstates$, to ensure that at least one action is chosen for each memory state:
\begin{displaymath}
\bigvee_{j \in \act}A_{mj}
\end{displaymath}

To ensure that the memory update function is well-defined, we introduce the following clause for each $m \in \memstates$, $a \in \act$ and $z \in \obs$. 
\begin{displaymath}
\bigvee_{m' \in \memstates}M_{m, z, a, m'} 
\end{displaymath}

The following clauses ensure that the $\{C_{i,m}\}$ variables will be assigned True, for all pairs $(i,m)$ that are reachable using the strategy.  
\begin{displaymath}
\neg C_{i,m} \vee \neg A_{m,a} \vee \neg M_{m, z,a, m'} \vee C_{j,m'}
\end{displaymath}
Such a clause is defined for each pair of memory-states $m,m' \in \memstates$, each pair of states $i,j \in \states$, each observation $z \in \obs$, and each action $a \in \act$, such that $\trans(i, a)(j) >0$ and $z = \obsmap(j)$.  

Clearly, the initial state and initial memory state are reachable. This is enforced by adding the single clause:
\begin{displaymath}
(C_{I,m_0})
\end{displaymath} 

We introduce the following unit clause for each $m \in \memstates$ and  $0\leq j \leq k$, which says that the goal state with any memory-state is reachable from the goal state and that memory-state, using a path of length at most $0$:
\begin{displaymath}
(P_{G,m,j})
\end{displaymath} 

Next, we define the following binary clause for each $i \in \states$ and $m \in \memstates$, so that if the (state, memory-state) pair $(i,m)$ is reachable, then the existence of a path from $(i,m)$ to the goal is enforced.  
\begin{displaymath}
\neg C_{i,m} \vee P_{i,m,k}
\end{displaymath}

Finally, we use the following constraints to define the value of the $P_{i,m,j}$ variables in terms of the chosen strategy.
\begin{equation*}
\begin{array}{l}
\displaystyle P_{i,m,j} \iff \\
\displaystyle \bigvee_{a \in \act}\left[A_{ma} \wedge \left(\bigvee_{\substack{m' \in \memstates, z \in \obs,\\
i' \in \states : \trans(i,a)(i') > 0\\
 \textrm{ and } \obsmap(i') = z}}\left[M_{m,z,a,m'} \wedge P_{i',m',j-1} \right]\right)\right] 
\end{array}
\end{equation*}

This constraint is defined for each $i\in \states$, $m \in \memstates$ and $1\leq j \leq k$. We use the standard Tseitin encoding to translate this formula to clauses.
The conjunction of all clauses defined above forms the CNF formula $\Phi_{k,\msize}$.

\begin{theorem}
If $\Phi_{k,\msize}$ is satisfiable then there is a winning, small-memory strategy of size $\msize$, and such a strategy is defined by the truth assignment to the $\{A_{ma}\}$ and $\{M_{m,z,a,m'}\}$ variables. If $\Phi_{k,\msize}$ is unsatisfiable, and $k \geq |\states|\cdot \msize$, then there is no small-memory strategy of size $\msize$ that is winning. 
\end{theorem}

The number of variables in $\Phi_{k,\msize}$ is $O(|\states|\cdot \msize \cdot k + \msize^2 \cdot |\obs|\cdot |\act|)$. The number of clauses is $O(|\states|^2\cdot \msize^2 \cdot |\obs|\cdot|\act|\cdot k)$. The number of actions, $|\act|$, and the number of observations, $|\obs|$, are usually constants. We also expect that the number of memory states, $\msize$, is small. Since $k \leq |\states|\cdot \msize$, the number of variables is quadratic and the number of clauses is cubic in the size of the state space, as for memoryless strategies.     

The comments in Remark~\ref{remark} also carry over to the small-memory case. In practice, we can often find a winning strategy with 
small values for $k$ and $\msize$ (see Section~\ref{sec:exp}). 

\begin{remark}
Our encoding can be naturally extended to search for deterministic strategies, for details see Appendix~\ref{sec:app:deterministic}.
\end{remark}

\section{Experimental Results}\label{sec:exp}
In this section we present our experimental results, which show that small-memory winning 
strategies do exist for several realistic POMDPs that arise in practice. 
Our experimental results clearly demonstrate the scalability of our SAT-based approach, 
which yields good performance even for POMDPs with large state spaces, 
where the previous explicit approach performs poorly.

We have implemented the encoding for small-memory strategies, described in 
Section~\ref{sec:encsmall}, as a small python program. 
We compare against the explicit graph-based algorithm 
presented in~\cite{CCGK14}. This is the state-of-the-art explicit POMDP solver 
for almost-sure reachability based on path-finding algorithms of~\cite{CH14} with a number of heuristics.
We used the SAT solver Minisat, version 2.2.0~\cite{ES04}.
The experiments were conducted on a Intel(R) Xeon(R) @ 3.50GHz 
with a 30 minute timeout.
We do not report the time taken to generate the encoding using our python 
script, because it runs in polynomial time, and more efficient implementations 
can easily be developed.
Also, we do not exploit incremental SAT in our 
experimental results (this will be part of future work).
We consider several POMDPs that are similar to well-known benchmarks. 
We generated several instances of each POMDP, 
of different sizes, in order to test the scalability of our algorithm.

\smallskip\noindent{\bf Hallway POMDPs.}
We considered a family of POMDP instances, inspired by the Hallway problem introduced in~\cite{LCK95} and used later in~\cite{S04,SS04,BG09,CCGK14}. 
In the Hallway POMDPs, a robot navigates on a rectangular grid. The grid has barriers where the robot cannot move, as well as trap locations that destroy the robot. The robot must reach a specified goal location. The robot has three actions: move forward, turn left, and turn right. The robot can see whether there are  barriers around its grid cell, so there are two observations (wall or no wall) 
for each direction. The actions may all fail, in which case the robot's state remains the same. The state is therefore comprised of the robot's location in the grid, and its orientation. Initially, the robot is randomly located somewhere within a designated subset of grid locations, and the robot is oriented to the south (the goal is also to the south). 
We generated several Hallway instances, of sizes shown in Table~\ref{table:hallway}. 
The runtimes for the SAT-based approach and the explicit approach are also given in the table. 
Timeouts (of 30 minutes) are indicated by ``-''. 
In all cases, the number of memory states required for there to be a winning strategy is $2$. 
Therefore, the runtimes reported for $\msize = 1$ correspond to the time required by the SAT solver 
to prove that $\Phi_{k,\msize}$ is unsatisfiable, while runs where $\msize=2$ resulted in 
the SAT solver finding a solution.  
We set $k$ to a sufficiently large value by inspection of the POMDP instance.  

\begin{table}[h!]
\begin{center}
\begin{tabular}[h]{l|l|l|l|l|l|l|l}
\toprule
Name & Grid &\!\# States\!&\!Explicit\!&    \multicolumn{4}{c}{Minisat (s)}   \\
& & & (s) &    \multicolumn{2}{c|}{UNSAT} & \multicolumn{2}{c}{SAT}   \\
     &      &           &   &  $k$ & $\msize=1$ & $k$ & $\msize=2$\\
\midrule
HW1 & $11\times 8$ & 3573 & 128.9 & 14 & 0.1 & 14 &0.6 \\
HW2 & $11\times 9$ & 4189 & - & 16 & 0.1 & 16 & 0.9 \\
HW3 & $11\times 10$ & 4981 & - & 18 &  0.2& 18 & 2.0 \\
HW4 & $15 \times 12$ & 9341 & - & 22 & 0.6& 22& 10.4 \\
HW5 & $19\times 14$ & 15245 & - & 30 & 2.0& 30 &81.9\\
HW6 & $23 \times 16$ & 22721 & - & 35 & 5.9& 35 & 244.6\\
HW7 & $27 \times 18$ & 31733 & - & 40 & 18.0& 40 & 635.7\\
HW8 & $29 \times 20$ & 39273 & - & 45 & 55.4& 45 &1157.1 \\ 
HW9 & $31 \times 22$ & 47581 & - & 50 & 127.9 & 50 & -\\   
\bottomrule
\end{tabular}
\end{center}
\caption{Results of the explicit algorithm and our SAT-based approach, on the Hallway instances.}
\label{table:hallway}
\end{table}

\smallskip\noindent{\bf Escape POMDPs.}
The problem is based on a case study published in~\cite{SCL15}, where the goal is 
to compute a strategy to control a robot in an uncertain environment.
Here, a robot is navigating on a square grid. There is an agent moving around the grid, and the robot must avoid being captured by the agent, forever. 
The robot has four actions: move north, move south, move east, move west. 
These actions have deterministic effects, i.e., they always succeed. 
The robot can observe whether or not there are barriers in each direction, and it can also observe the position of the agent 
if the agent is currently on an adjacent cell. The agent moves randomly. 
We generated several instances of the Escape POMDPs, of sizes shown in Table~\ref{table:escape}. 
The runtimes for the SAT-based approach and the explicit approach are also given in the table, with timeouts indicated by ``-''. The number of memory states was set to $\msize = 5$, which is sufficient for there to be a small-memory winning strategy.
For these POMDPs, there is always a path directly to the goal state, so setting $k=2$ was sufficient to find a winning strategy. In order to prove that there is no smaller winning strategy, we increased $k$ to $8 = 2\times \msize$, where $\msize=4$. The runtimes for the resulting unsatisfiable formulas are also shown in Table~\ref{table:escape}.

\begin{table}[h!]
\begin{center}
\begin{tabular}[h]{l|l|l|l|l|l|l|l}
\toprule
Name & Grid & \# States & Explicit (s) & \multicolumn{4}{c}{Minisat (s)} \\
& & &  &    \multicolumn{2}{c|}{UNSAT} & \multicolumn{2}{c}{SAT}   \\
     &   &         &   &  $k$ & $\msize=4$ & $k$ & $\msize=5$ \\
\midrule
Escape3 & $3\times 3$ & 84 & 0.4 & 8 & 0.4 & 2 &0.2\\
Escape4 & $4\times 4$ & 259 & 0.9 & 8 & 1.56 & 2 & 1.0 \\
Escape5 & $5 \times 5$ & 628 & 6.8 & 8 & 5.0 & 2 & 3.3 \\
Escape6 & $6\times 6$ & 1299 & 20.9 & 8 & 15.8 & 2 & 9.0 \\
Escape7 & $7\times 7$ & 2404 & 89.2 &8 & 36.2 & 2 & 19.5 \\
Escape8 & $8\times 8$ & 4099 & 238.6 & 8 & 63.4 & 2 & 47.1 \\
Escape9 & $9 \times 9$ & 6564 & 688.6 & 8 & 113.5 & 2 & 60.2 \\
Escape10 & $10\times 10$ & 10003 & - & 8 & 212.6 & 2 &113.1\\
Escape11 & $11\times 11$ & 14644 & - & 8 & 303.3 & 2 & 210.4\\
Escape12 & $12 \times 12$ & 20739 & - & 8 & 535.4 & 2 & 505.1\\
\bottomrule
\end{tabular}
\end{center}

\caption{Results of the explicit algorithm and our SAT-based approach, on the Escape instances.}
\label{table:escape}
\end{table}

\smallskip\noindent{\bf RockSample POMDPs.}
We consider a variant of the RockSample problem introduced in~\cite{SS04} and used later in~\cite{BG09,CCGK14}. 
The RockSample instances model rover science exploration. The positions of the rover and the rocks are known, 
but only some of the rocks have a scientific value; we will call these rocks good. 
The type of the rock is not known to the rover, until the rock site is visited.
Whenever a bad rock is sampled the rover is destroyed and a losing absorbing state is reached. 
If a sampled rock is sampled for the second time, then with probability $0.5$ the action has no effect. 
With the remaining probability the sample is destroyed and the rock needs to be sampled one more time.
An instance of the RockSample problem is parametrized with a parameter $[n]$: $n$ is the number 
of rocks on a grid of size $3 \times 3$. The goal of the rover is to obtain two samples of good rocks. 
In this problem we have set $\mu=2$ and $k=8$, which is sufficient to find a winning strategy for each instance.
However, memoryless strategies are not sufficient as in some situations sampling is prohibited 
whereas in other situations it is required (hence we did not consider $\mu=1$). 
The results are presented in Table~\ref{table:rocksample}.

\begin{table}[h!]
\centering
\begin{tabular}[h!]{l|l|l|l}
\toprule
Name & \# States & Explicit (s) & Minisat (s) \\
\midrule
RS[4]  & 351 & 0.4 & 0.06 \\
RS[5]  & 909 & 1.6 & 0.24 \\
RS[6]  & 2187 & 3.4 & 0.67 \\
RS[7]  & 5049 & 14.3 & 1.58 \\
RS[8]  & 11367 & 50.6 & 4.59 \\
RS[9]  & 25173 & 197.3 & 79.1 \\
\bottomrule
\end{tabular}
\caption{Results of the explicit algorithm and our SAT-based approach, on the RockSample instances.}
\label{table:rocksample}
\end{table} 

\begin{remark}
In the unsatisfiable (UNSAT) results of the Hallway and Escape POMDPs, we have computed, based on the diameter of the underlying graph and the number of memory elements $\mu$, a sufficiently large $k$ to disprove the existence of an almost-sure winning strategy of the considered memory size. It follows, that there is no memoryless strategy for the Hallway POMDPs, and no almost-sure winning
strategy for the Escape POMDPs, that uses only $4$ memory elements.
\end{remark}

\noindent\textbf{Memory requirements.} 
In all runs of the Minisat solver, at most $5.6$ GB of memory was used. 
The runs of the explicit solver consumed around $30$ GB of memory 
at the timeout.

\section{Conclusion and Future Work}
In this work we present the first symbolic SAT-based algorithm for 
almost-sure reachability in POMDPs. 
We have illustrated  
that the symbolic algorithm significantly outperforms the explicit algorithm, on a number of examples similar to problems from the literature. 
In future work we plan to investigate the possibilities of incremental SAT solving.
Incremental SAT solvers can be beneficial in two ways: 
First, they may improve the efficiency of algorithms to 
find the \emph{smallest} almost-sure 
winning strategy. Such an approach can be built on top of our encoding. Second, incremental SAT solving could help in the case that the original POMDP is modified slightly, in order to efficiently solve the updated SAT instance. Investigating the practical impact of incremental SAT solvers for POMDPs is the subject of future work.

\bibliographystyle{plain}
\bibliography{pomdp}

\newpage
\appendix

\section{Appendix: Detailed comparison of qualitative analysis and contingent planning.}
\label{sec:app:cont}
We present examples that distinguish almost-sure winning from contingent planning.
We first explain the conceptual difference and then illustrate the difference with examples. 
In the contingent planning setting it is required that \emph{all} paths reach the 
goal state. 
In other words, contingent planning treats the probabilistic choice as an adversarial 
choice.
In almost-sure winning, although it is true that the precise probabilities do not 
matter, it is still different than treating the probabilistic choice as adversarial.
We first illustrate the difference with examples of Markov chains.

\begin{example}[Markov chains.]
In Figure~\ref{fig:mc1} we depict a Markov Chain $M_1$ (which is a perfect-information 
MDP with a single action) with two states: 
the initial state $s_0$ and the goal state $G$. 
The probabilistic transition function in state $s_0$ selects the next state to be $s_0$ 
with probability $\frac{1}{2}$, and the goal state $G$ with the remaining probability $\frac{1}{2}$.
In this example, the probability to reach $G$ in $n$ steps is $\sum_{i=1}^{n}(\frac{1}{2})^i$. 
Since we consider the infinite-horizon setting, by taking the limit of $n$ to $\infty$ we obtain that 
the probability of eventually reaching the goal state $G$ is $1$, i.e,  
$$\lim_{n \rightarrow \infty} \sum_{i=1}^{n}\left(\frac{1}{2}\right)^i = \lim_{n \rightarrow \infty} (1 - \frac{1}{2^n}) =1$$
Hence in this example, the goal state is reached almost-surely (with probability $1$).
However, in the contingent planning setting, there is no plan, since there exists a path 
that stays in $s_0$ forever (namely, $s_0^\omega$) that does not reach the goal state $G$.
Hence the answers are different: the answer to almost-sure winning is YES, whereas the answer to contingent planning
is NO.
Note that in the Markov chain example, if the probabilities change from $(\frac{1}{2},\frac{1}{2})$
to $(\frac{2}{3},\frac{1}{3})$ or $(\frac{3}{4},\frac{1}{4})$ the answer to the almost-sure winning
still remains same (the probability to reach within $n$ steps changes to $(1-(\frac{2}{3})^n)$ and
$(1-(\frac{3}{4})^n)$, respectively, however the limits are still~1).
For almost-sure winning the precise probability values of transitions do not matter,
nevertheless this is still different from treating the probabilistic choices
as adversarial choice (as considered in contingent planning). 
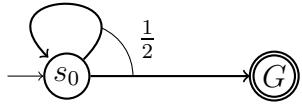
\begin{figure}[h]
\centering  
\begin{tikzpicture}
  \tikzstyle{every node}=[font=\large]
  \node[State,initial,initial text=,solid] (s) {$s_0$};
  \node[State,right of=s,xshift = 50, accepting,solid] (g) {$G$};

  \draw[->,thick]{
	(s) edge[]  (g)
	(s) edge[loop]  (s)
  };
  \draw ($(s) +(25pt, 0pt)$) arc (0:66:0.7cm);
  \node[right of=s,xshift=2, yshift=14] {$\frac{1}{2}$};
\end{tikzpicture}
\label{fig:mc1}
\caption{Markov Chain $M_1$}
\end{figure}

Next we consider the example $M_2$ shown in Figure~\ref{fig:mc2}, where from $s_0$, the next state is 
one of the three states $L$, $G$, and $s_0$, with probabilities $\frac{1}{3}$ for each.
Here, the answer to both the almost-sure winning and contingent planning questions is NO.
However, the path $s_0^\omega$ that stays in $s_0$ forever is a witness to show that the answer
to the contingent planning problem is NO, but the same witness is not valid for almost-sure winning.
In other words, even when the answers to almost-sure winning and contingent planning are the same,
a witness to show that the answer for contingent planning is NO, is not necessarily a 
witness to show the answer to almost-sure winning is NO.
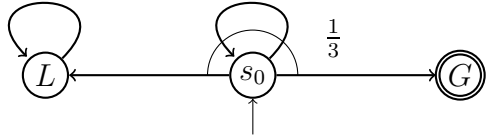
\begin{figure}[h]
\centering  
\begin{tikzpicture}
  \tikzstyle{every node}=[font=\large]
  \node[State,initial below,initial text=,solid] (s) {$s_0$};
  \node[State,right of=s,xshift = 50, accepting,solid] (g) {$G$};
  \node[State,left of=s,xshift = - 50, solid] (l) {$L$};

  \draw[->,thick]{
	(s) edge[]  (g)
	(s) edge[]  (l)
	(s) edge[loop]  (s)
	(l) edge[loop]  (l)
  };
  \draw ($(s) +(17pt, 0pt)$) arc (0:180:0.6cm);
  \node[right of=s,xshift=2, yshift=14] {$\frac{1}{3}$};
\end{tikzpicture}
\caption{Markov Chain $M_2$}
\label{fig:mc2}
\end{figure}
\end{example}

\begin{example}[Perfect Information MDPs.]
We now illustrate the situation in a perfect-information MDP. Note that the situation would be only more complicated in the general POMDP setting.
Consider the MDP shown in Figure~\ref{fig:mdp1}.
The initial state is $s_0$, and there are two actions available at $s_0$, to either go to state $V$ 
with action $a$ or to state $U$ with action $b$. 
In state $V$, with probability $\frac{1}{3}$ the next state is $U$, $s_0,$ or $G$
(irrespective of the actions).
In state $U$, with probability $\frac{1}{2}$ the next state is $U$ or $s_0$
(irrespective of the actions).
In this example, a strategy that always chooses action $a$ at $s_0$ is an 
almost-sure winning strategy, but there is no strategy that ensures that the answer to the 
contingent planning problem is YES.
Also note that in this example, a strategy that always chooses $b$ at $s_0$
is not an almost-sure winning strategy.
Thus even when almost-sure winning strategies  exist, not all strategies are 
almost-sure winning.

\begin{figure}[h]
\centering  

\begin{tikzpicture}
  \tikzstyle{every node}=[font=\large]
  \node[State,initial below,initial text=,solid] (s) {$s_0$};
  \node[State,right of=s,xshift = 50, solid] (v) {$V$};
  \node[State,right of=v,xshift = 50, accepting,solid] (g) {$G$};
  \node[State,left of=s,xshift = - 50, solid] (u) {$U$};

  \draw[->,thick]{
	(u) edge[bend right=20]  (s)
	(s) edge[bend right=20] node[above] {$a$} (v)
	(v) edge[]  (g)
	(u) edge[loop]  (u)
	(s) edge[bend right=40] node[above] {$b$} (u)
	(v) edge[bend right=40] (s)
	(v) edge[bend left=35] (u)

  };
  \draw ($(v) +(-10pt, -8pt)$) arc (-144:144:0.5cm);
  \node[right of=v,xshift=3, yshift=14] {$\frac{1}{3}, a,b$};
  \draw ($(u) +(19pt, -6pt)$) arc (-30:65:0.5cm);
  \node[right of=u,xshift=7, yshift=5] {$\frac{1}{2}, a,b$};
\end{tikzpicture}
\caption{MDP $M_3$}
\label{fig:mdp1}
\end{figure}
\end{example}

In summary, the above examples illustrate the following: 
\begin{compactenum}
\item The answer to the almost-sure winning question on a given input can be very different from contingent 
planning, even in very special cases of POMDPs, namely, in 
perfect-information Markov chains.
\item Even in cases when the answer to the almost-sure winning and contingent planning questions is
the same, not every witness for the contingent planning problem is a valid witness for the almost-sure 
winning problem (even for Markov chains).
\item In case of perfect-information MDPs, the almost-sure winning strategy 
construction can be quite involved, and different from contingent 
planning.
\end{compactenum}
The key difference of the two setting is as follows: in the contingent planning 
since the requirement is for all paths, this effectively means treating the 
probabilistic choice as an adversary. 
For almost-sure winning, if a probabilistic choice is available infinitely often,
then it must be chosen infinitely often.
Note that almost-sure winning is the classical probability theory counterpart of 
almost-sure convergence, which is the strongest probabilistic guarantee, yet it
does not require convergence for all points.

\section{Appendix: Detailed comparison of qualitative analysis and strong cyclic planning.}
\label{sec:app:cont2}

The qualitative analysis problem is equivalent to the strong cyclic planning problem.
The strong cyclic problem was studied in the perfect information setting in~\cite{CPRT03} 
and later extended to the partial information setting in~\cite{BCP06}. 
However, there are two crucial differences of our work wrt~\cite{BCP06}: 
\begin{enumerate}
\item We consider the problem of finding small strategies as compared to general strategies. 
We show that our problem is NP-complete. In contrast, it is known that the qualitative analysis 
problem for POMDPs with general strategies is EXPTIME-complete~\cite{CDH10a,BGB12}. 
Thus the strong cyclic planning with general strategies considered in~\cite{BCP06} 
is also EXPTIME-complete, whereas we establish that our problem is NP-complete.
Thus there is a significant difference in the complexity of the problem finding 
small strategies as compared to general strategies.

\item The work of~\cite{BCP06} presents a BDD-based implementation, whereas we present a 
SAT-based implementation. 
Note that since~\cite{BCP06} considers an EXPTIME-complete problem in general there is no 
efficient (polynomial-time) reduction to SAT.
In contrast not only we show that our problem is NP-complete, we present an efficient 
(cubic for constant-size memory) reduction to SAT.
\end{enumerate}
To the best of our knowledge, there is no publicly available implementation for 
strong cyclic planning under partial observation.

\begin{table}[h!]
\begin{tabular}[h]{c|c|c}
\toprule
& MDP & POMDP \\
\midrule
\multirow{2}{*}{Strong planning} & SAT-based algorithm & MBP BDD-based\\
& \cite{BEZ05} & \cite{CPRT03}\\[0.2em]
\midrule
\multirow{4}{*}{Strong cyclic planning} & SAT-based algorithm & BDD-based\\
& \cite{BEZ05} & \cite{BCP06}\\[0.8em]
& MBP BDD-based & \textbf{SAT-based for small strategies}\\
& \cite{CPRT03} & \textbf{Theorem~\ref{thm:small}}\\
\bottomrule
\end{tabular}
\caption{Comparison of existing algorithms}
\label{table:algorithms}
\end{table}

\smallskip\noindent{\em Significance of our result.}
Finally, the equivalence of strong cyclic planning and qualitative analysis of
POMDPs implies a greater significance of our result. 
First, our results become applicable also for strong cyclic planning. 
Second, our approach gives a way to compute small strategies (if they exist)
for strong cyclic planning. 
Finally, we present an an efficient SAT-based implementation to
obtain small strategies in strong cyclic planning.
Previous works consider BDD-based approach to compute general strategies.
Developing fast SAT-solvers (or incremental SAT-solvers) is an active 
research area, and our results imply that faster solvers for SAT can then
be used both for qualitative analysis of POMDPs as well as for 
strong cyclic planning, for computing small strategies when they exist.
In Table~\ref{table:algorithms} we present the comparison of the existing approaches
for strong planning and strong cyclic planning for MDPs (perfect-information 
setting) as well as POMDPs.
Note that as described in the previous section, the strong planning problem
is different from the qualitative analysis of POMDPs (or strong cyclic planning).
In the perfect-information setting there exists SAT-based solvers both for 
strong planning as well as strong cyclic planning, whereas for general 
strategies in POMDPs there exists no SAT-based implementation.
We present the first SAT-based implementation to compute small strategies
for strong cyclic planning in the partial information setting.

\section{Appendix: Deterministic Strategies}
\label{sec:app:deterministic}
In this part we present a simple extension of our encoding that handles the case for deterministic strategies.

\smallskip\noindent \textbf{Deterministic Strategies.}
A strategy with memory $\straa = (\straa_u, \straa_n, \memstates, m_0)$ is \emph{deterministic} if both functions $\straa_n$ and $\straa_u$
assign only Dirac probability distributions and can be written as:

\begin{itemize}
\item The \emph{action selection function} is of type $\straa_n : \memstates \rightarrow \act$.
\item The \emph{memory update function} is of type $\straa_u : \memstates \times \obs \times \act \rightarrow \memstates$.
\end{itemize}

We will present the modification only for the more complicated case of small-memory strategies, the modifications for the memoryless case are analogous.

\smallskip\noindent \emph{Next-action Selection Function.}
The part of the encoding that codes for the \emph{next-action selection function} $\straa_n$ is:

\begin{displaymath}
\bigvee_{j \in \act}A_{mj}
\end{displaymath}

It ensures for every $m \in M$, that at least one variable $A_{mj}$ is set to true, i.e., at least one action is chosen. In a deterministic strategy the requirements are stronger, it is required that exactly one action is chosen. This can be enforced by adding the following clause for every memory element $m$, and two distinct actions $i,j$:

\begin{displaymath}
A_{mi} \oplus A_{mj}
\end{displaymath}

\smallskip\noindent \emph{Memory Update Function.}
The part of the encoding that codes for the \emph{memory update function} $\straa_u$ is:

\begin{displaymath}
\bigvee_{m' \in \memstates}M_{m, z, a, m'} 
\end{displaymath}

As in the previous case, it is sufficient to add the following clause for every memory element $m$, observation $z$, action $a$, and two distinct memory element $m'$ and $m"$:

\begin{displaymath}
M_{m, z, a, m'} \oplus M_{m, z, a, m"}
\end{displaymath}

\end{document}